\documentclass[conference]{IEEEtran}

\usepackage{cite}
\usepackage{amsmath,amssymb,amsfonts}
\usepackage{algorithmic}
\usepackage{graphicx}
\usepackage{textcomp}
\usepackage{xcolor}

\def\BibTeX{{\rm B\kern-.05em{\sc i\kern-.025em b}\kern-.08em
    T\kern-.1667em\lower.7ex\hbox{E}\kern-.125emX}}
    
\RequirePackage[textsize=scriptsize]{todonotes}
\RequirePackage{enumitem} \setlist{nolistsep}
\RequirePackage[normalem]{ulem}
\RequirePackage{soul}
\usepackage{multirow}
\RequirePackage{flushend}
\RequirePackage{soul}

\DeclareSymbolFont{symbolsC}{U}{txsyc}{m}{n}
\SetSymbolFont{symbolsC}{bold}{U}{txsyc}{bx}{n}
\DeclareFontSubstitution{U}{txsyc}{m}{n}
\DeclareSymbolFont{largesymbolsA}{U}{txexa}{m}{n}
\SetSymbolFont{largesymbolsA}{bold}{U}{txexa}{bx}{n}
\DeclareFontSubstitution{U}{txexa}{m}{n}
\DeclareMathDelimiter{\llbracket}{\mathopen}{symbolsC}{126}{largesymbolsA}{18}
\DeclareMathDelimiter{\rrbracket}{\mathclose}{symbolsC}{127}{largesymbolsA}{19}

\def\R{\mathbb{R}}
\DeclareMathOperator{\softmax}{SoftMax}
\def\LSTMlr{\ensuremath{\text{LSTM}_{\textsc{lr}}}}
\def\LSTMrl{\ensuremath{\text{LSTM}_{\textsc{rl}}}}
\def\Wtdlstm{\ensuremath{W_{\textsc{tdlstm}}}}
\def\GRUauxLR{\ensuremath{\text{GRU}^{\textsc{aux}}_{\textsc{lr}}}}
\def\GRUauxRL{\ensuremath{\text{GRU}^{\textsc{aux}}_{\textsc{rl}}}}
\def\Waux{\ensuremath{W_\textsc{aux}}}
\def\GRUmainLR{\ensuremath{\text{GRU}^{\textsc{main}}_{\textsc{lr}}}}
\def\GRUmainRL{\ensuremath{\text{GRU}^{\textsc{main}}_{\textsc{rl}}}}
\def\Wmain{\ensuremath{W_{\textsc{main}}}}

\usepackage{booktabs} % For formal tables
\usepackage{url}
\usepackage[numbers]{natbib}

\begin{document}

\title{Multi-task Learning for Target-Dependent Sentiment Classification}

\if 0
\author{\IEEEauthorblockN{1\textsuperscript{st} Given Name Surname}
\IEEEauthorblockA{\textit{dept. name of organization (of Aff.)} \\
\textit{name of organization (of Aff.)}\\
City, Country \\
email address}
\and
\IEEEauthorblockN{2\textsuperscript{nd} Given Name Surname}
\IEEEauthorblockA{\textit{dept. name of organization (of Aff.)} \\
\textit{name of organization (of Aff.)}\\
City, Country \\
email address}
\and
\IEEEauthorblockN{3\textsuperscript{rd} Given Name Surname}
\IEEEauthorblockA{\textit{dept. name of organization (of Aff.)} \\
\textit{name of organization (of Aff.)}\\
City, Country \\
email address}
\and
\IEEEauthorblockN{4\textsuperscript{th} Given Name Surname}
\IEEEauthorblockA{\textit{dept. name of organization (of Aff.)} \\
\textit{name of organization (of Aff.)}\\
City, Country \\
email address}
\and
\IEEEauthorblockN{5\textsuperscript{th} Given Name Surname}
\IEEEauthorblockA{\textit{dept. name of organization (of Aff.)} \\
\textit{name of organization (of Aff.)}\\
City, Country \\
email address}
\and
\IEEEauthorblockN{6\textsuperscript{th} Given Name Surname}
\IEEEauthorblockA{\textit{dept. name of organization (of Aff.)} \\
\textit{name of organization (of Aff.)}\\
City, Country \\
email address}
}

\fi

\maketitle

\begin{abstract}
Detecting and aggregating sentiments toward people, organizations, and events expressed in unstructured social media have become critical text mining operations.  Early systems detected sentiments over whole passages, whereas more recently, target-specific sentiments have been of greater interest.  In this paper, we present MTTDSC, a multi-task target-dependent sentiment classification system that is informed by feature representation learnt for the related auxiliary task of passage-level sentiment classification.  The auxiliary task uses a gated recurrent unit (GRU) and pools GRU states, followed by an auxiliary fully-connected layer that outputs passage-level predictions.  In the main task, these GRUs contribute auxiliary per-token representations over and above word embeddings.  The main task has its own, separate GRUs.  The auxiliary and main GRUs send their states to a different fully connected layer, trained for the main task.  Extensive experiments using two auxiliary datasets and three benchmark datasets (of which one is new, introduced by us) for the main task demonstrate that MTTDSC outperforms state-of-the-art baselines.  Using word-level sensitivity analysis, we present anecdotal evidence that prior systems can make incorrect target-specific predictions because they miss sentiments expressed by words independent of target.  Thus, our superior accuracy is also backed by a more interpretable model.
\end{abstract}


% \begin{document}
\title{Multi-task Learning for Target-dependent \\ Sentiment Classification}
% \title{Target-dependent Sentiment Classification and Anomaly Detection \\ Using Multi-task Learning}

\author{Divam Gupta$^1$, Kushagra Singh$^2$, Soumen Chakrabarti$^4$, Tanmoy Chakraborty$^3$}
\institute{$^{1,2,3}$IIIT Delhi, India , $^{4}$IIT Bombay, India\\
\{$^1$divam14038,$^2$kushagra14056,$^3$tanmoy\}@iiitd.ac.in;\ $^4$soumen@cse.iitb.ac.in
}
\authorrunning{Gupta et al.} % abbreviated author list (for running head)
%
%%%% list of authors for the TOC (use if author list has to be modified)
%\tocauthor{Ivar Ekeland, Roger Temam, Jeffrey Dean, David Grove,
%Craig Chambers, Kim B. Bruce, and Elisa Bertino}
%

\maketitle              % typeset the title of the contribution

\begin{abstract}
Detecting and aggregating sentiments toward people, organizations, and events expressed in unstructured social media have become critical text mining operations. 
Early systems detected sentiments over whole passages, whereas more recently, target-specific sentiments have been of greater interest. 
In this paper, we present MTTDSC, a multi-task target-dependent sentiment classification system that is informed by feature representation learnt for the related auxiliary task of passage-level sentiment classification.  The auxiliary task uses a gated recurrent unit (GRU) and pools GRU states, followed by an auxiliary fully-connected layer that outputs passage-level predictions.  In the main task, these GRUs contribute auxiliary per-token representations over and above word embeddings.  The main task has its own, separate GRUs.  The auxiliary and main GRUs send their states to a different fully connected layer, trained for the main task.  Extensive experiments using two auxiliary datasets and three benchmark datasets (of which one is new, introduced by us) for the main task demonstrate that MTTDSC outperforms state-of-the-art baselines.  Using word-level sensitivity analysis, we present anecdotal evidence that prior systems can make incorrect target-specific predictions because they miss sentiments expressed by words independent of target.  
%Apart from such intrinsic evaluation, we also evaluate MTTDSC in the extrinsic task of detecting anomalous social media users whose sentiment toward related targeted entities are inconsistent.
\end{abstract}

\section{Introduction}
\label{sec:Intro}

As the volume of news, blogs \citep{GodboleSS2007NewsBlogSentiment},
and social media \citep{PakP2010TwitterSentiment}
far outstrips what an individual can consume, sentiment classification (\textbf{SC}) \citep{PangL+2008OpinionSentiment,Liu2012SentimentOpinion} 
has become a powerful tool for understanding emotions toward politicians, celebrities, products, governance decisions, etc. Of particular interest is to identify sentiments expressed toward specific entities, i.e., {\em target dependent sentiment classification} (\textbf{TDSC}).  Recent years have witnessed many TDSC approaches \citep{WilsonWH2005ContextualPolarity, 
TengVZ2016LexiconSentiment, WangLZP2017TDParse} with increasing sophistication and accuracy.

Possibly because research on passage-level and target-dependent sentiment classification were separated in time by the dramatic emergence of deep learning, TDSC systems predominantly use recurrent neural networks (RNNs) %\citep{HochreiterS1997LSTM, ChungGCB2014GRUvsLSTM} 
and borrow little from passage-level tasks and trained models.  From the perspective of curriculum learning \citep{BengioLCW2009Curriculum}, this seems suboptimal: representation borrowed from passage-level SC should inform TDSC well.  Moreover, whole-passage labeling entails considerably lighter cognitive burden than target-specific labeling.  As a result, whole-passage gold labels can be collected at larger volumes.

In this paper, we present MTTDSC, {\bf M}ulti-{\bf T}ask {\bf T}arget {\bf D}ependent {\bf S}entiment {\bf C}lass\-ifier\footnote{MTTDSC code and datasets are available at \url{https://github.com/divamgupta/mttdsc } }, a novel multi-task learning (MTL) system that uses passage-level SC as an auxiliary task and TDSC as the main task. 
MTL has shown significant improvements in many fields of Natural Language Processing and Computer Vision.  In basic (`naive') MTL, we jointly train multiple models for multiple tasks with some shared parameters, usually in network layers closest to the inputs \citep{MaurerPR2016MTL}, resulting in shared input representation learning.  Symmetric, uncontrolled sharing can be detrimental to some tasks.  %Finer control over sharing is achieved by Sluice networks \citep{ruder2017learning}, but at the cost of more training data.
% We show that neither naive nor Sluice MTL matches the performance of MTTDSC's inter-task sharing structure, which is customized to the TDSC application.

In MTTDSC, the auxiliary SC task uses bidirectional GRUs, whose states are pooled over positions to make whole-passage predictions.  This sensitizes the auxiliary GRU to target-independent expressions of sentiments in words.  The main TDSC task combines the auxiliary GRUs with its own target-specific GRUs.  The two tasks are jointly trained.  If passages with both global and target-specific labels are available, they can be shared between the tasks.  Otherwise, the two tasks can also be trained on disjoint passages.  MTTDSC can be interpreted as a form of task-level curriculum learning \citep{BengioLCW2009Curriculum}, where the simpler whole-passage SC task learns to identify sentiments latent in word vectors, which then assists the more challenging TDSC task.  Static sentiment lexicons, such as SentiWordNet \citep{AndreaS2006SentiWordNet}, are often inadequate for dealing with informal media.

Using two standard datasets, as well as one new dataset we introduce for the main task, we establish superiority of MTTDSC over several state-of-the-art approaches.  While improved accuracy from additional training data may seem unsurprising from a learning perspective, we show that beneficial integration of the auxiliary task and data is nontrivial.  Simpler multi-task approaches \citep{MaurerPR2016MTL}, where a common feature extraction network is used for jointly training on multiple tasks, perform poorly.  We also use word-level sensitivity tests to obtain anecdotal evidence that direct TDSC approaches (that do not borrow from whole-passage SC models) make target-specific prediction errors because they misclassify the (target independent) sentiments expressed by words.  Thus, MTTDSC also provides a more interpretable model, apart from accuracy gains.

% Over and above intrinsic evaluation of MTTDSC's accuracy at the SC task, we propose and evaluate mechanisms to use MTTDSC to detect anomalous users whose sentiment toward related entities are inconsistent.  MTTDSC performs well at this extrinsic evaluation as well.  

The contributions of our work are summarized as follows:
\begin{itemize}
\item MTTDSC, a {\bfseries novel neural MTL architecture designed specifically for TDSC}. We show the superiority of our model and also compare it with other state-of-the-art models of TDSC and multi-task learning. 
\item A {\bfseries new dataset} for target dependent sentiment classification which is better for real world analysis on social media data.
\item Thorough {\bfseries investigation of the reasons behind the success} of MTTDSC.  In particular, we show that current models fail to capture many emotive words owing to insufficient training data.
\end{itemize}

\section{Related work}
\label{sec:Related}

\textbf{\underline{Target dependent sentiment classification}:}
An input text passage is a sequence of words  $w_1, w_2, \ldots, w_N$.  We use `tweet' and  `passage' interchangeably, given the preponderance of social media in TDSC applications.  One word position or contiguous span is identified as a \emph{target}.  For simplicity, we will assume compounds like New\_York to be pre-fused  and consider a target as a single word position.  A passage may have one or more targets.  In gold labeled instances, the target is associated with one of three labels $\{-1,0,+1\}$, corresponding to negative, neutral, and positive sentiments respectively.  The passage-level task has a label associated with the whole passage \citep{Kim2014ConvnetSentence}, rather than a specific target position.  E.g., in the tweet ``I love listening to electronic music, however artists like \uwave{Avici} \& \uwave{Tiesta} copy it from others'', the overall sentiment is positive but the sentiments associated with both targets `Avici' and `Tiesta' are negative.

%Most approaches to TDSC collect word features around the target entity mention.  
TDLSTM \citep{TangQFL2015TDLSTM} and TDParse \citep{WangLZP2017TDParse} divide the sentence into left context, right context and the target entity, and then combine their features.  TDLSTM uses a left-to-right \LSTMlr\
on the context left of the target ($w_1,\ldots,w_i$),
a right-to-left \LSTMrl\ on the context right of the target ($w_i,\ldots,w_N$),
and a fully connected layer $\Wtdlstm$ that combines signals from
\LSTMlr\ and \LSTMrl.  If LSTM state vectors are in $\R^D$,
then\footnote{We elide possible scalar offsets in sigmoids and softmaxes for simplicity throughout the paper.} $\Wtdlstm \in \R^{2D\times 3}$.  Given a tweet and target position $i$,
\LSTMlr\ is applied to (pre-trained and pinned) input
word embeddings of $w_1,\ldots,w_i$, 
obtaining state vectors $\LSTMlr[1],\ldots,\LSTMlr[i]$.
Similarly, \LSTMrl\ is applied to (word embeddings of)
$w_i,\ldots,w_N$, obtaining state vectors
$\LSTMrl[i],\ldots,\LSTMrl[N]$.
The output probability vector in $\Delta^3$ is
\begin{align}
\softmax\Bigl(\bigl[\LSTMlr[i], \LSTMrl[i]\bigr]\,\Wtdlstm\Bigr),
\end{align}
where $\Delta^3$ is a 3-class multinomial distribution over $\{-1,0,1\}$, obtained
from the softmax.  Standard cross-entropy against the one-hot gold label is used for training.  TCLSTM \citep{TangQFL2015TDLSTM} is a slight modification to TDLSTM, where the authors also concatenated the embedding of the target entity with each token in the given sentence. They showed that TCLSTM has a slight improvement over TDLSTM.

By pooling embeddings of words appearing on dependency paths leading to the target position, TDParse \citep{WangLZP2017TDParse} improves further on TDLSTM accuracy.  More details of these systems are described in Section~\ref{sec:MethodsAndPerfs}, along with their performance. The major problem in TDParse is the inability to learn compositions of words. TDParse usually fails for the sentences containing a polar word which is not related to the entity.

The ``naive segmentation'' (Naive-Seg) model of \citet{WangLZP2017TDParse} concatenates word embeddings of left context, right context and sub sentences of the tweets. Various pooling functions are used to combine them and an SVM is used for labeling. Naive-Seg+ extends Naive-Seg by using sentiment lexicon based features. TDParse extends Naive-Seg by using dependency parse paths incident on the target entity to collect words whose embeddings are then pooled.  TDParse+ further extends TDParse by using sentiment lexicon \citep{AndreaS2006SentiWordNet} based features. TDParse+ beats TDLSTM largely because of carefully engineered features (including SentiWordNet based features), but may not generalize to diverse datasets.

Pooling word embeddings over dependency paths may not capture complex compositional semantics.  Given enough training data, TDLSTM should capture complex compositional semantics.  But in practice, neural sequence models start with word vectors that were not customized for sentiment detection, and then get limited training data.\\

\noindent\textbf{\underline{Multi-task learning}:}
Multi-task learning has been used in many applications related to NLP. 
\citet{peng2017deep} showed improved results in semantic dependency parsing be learning three semantic dependency
graph formalisms.  
% \citet{guo2016exploiting} jointly trained a semantic parsing model on multi-typed treebanks.
\citet{choi2017coarse} improved the performance on question answering by jointly training answer generation and answer retrieval model. 
Sluice networks proposed by \citet{ruder2017learning} claims to be a generalized model which could learn to control the sharing of information between different task models. 
Sluice networks do not perform well for TDSC, as the sharing of information happens at all positions of the sentence.
On the other hand, our model forces the auxiliary task to learn feature representation at all positions and share them at the appropriate locations with the main task.  

\section{MTTDSC: A multi-task approach to TDSC}
\label{sec:OurMethod}

Recurrent models for TDSC have to solve two challenging problems in one shot: identify sentiment-bearing words in the passage, and use influences between hidden states to connect those sentiments to the target.  A typical TDSC system attempts to do this as a self-contained learner, without representation support from an auxiliary learner solving the simpler task of whole-passage SC.  We present anecdotes in Section~\ref{sec:anecdotes} that reveal the limitations of such approaches.  In response, we propose a multi-task learning (MTL) approach called MTTDSC.  Representations trained for the auxiliary task (Section~\ref{sec:AuxTask}) inform the main task (Section~\ref{sec:MainTask}).  The combined loss objective is described in Section~\ref{sec:TwoTasks} and implementation details are presented in Section~\ref{sec:SystemDetails}.

Our MTL framework is significantly different from traditional ones.  In particular, we do not require auxiliary and main task gold labels to be available on the same instances.  (This makes it easier to collect larger volumes of auxiliary labeled data.)  As a result, in standard MTL, attempts to improve auxiliary task performance interferes with the learning of feature representations that are important for the main task.  To solve this problem, we use separate RNNs for the two tasks, the output of the auxiliary RNN acting as additional features to the main model.  This ensures that the gradients from the auxiliary task loss do not unduly interfere with the weights of the main task RNN.

\begin{figure*}[th]
\centering  \includegraphics[width=1.02\textwidth]{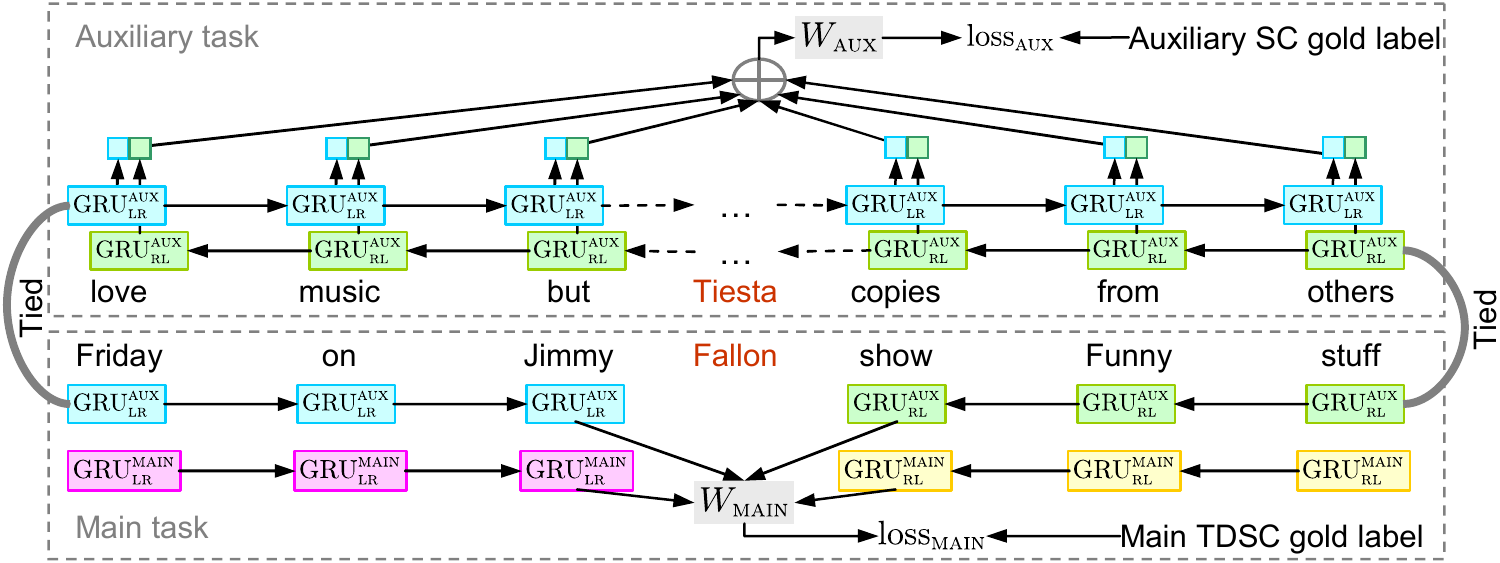}
\caption{MTTDSC network architecture.
Passage-level gold labels are used to compute loss in the upper auxiliary network. Target-level gold labels are used to compute loss in the lower main network. These are coupled through tied parameters in auxiliary GRUs.  The main task uses another set of task-specific GRUs.}
\label{fig:MTTDSCnetwork0}
\end{figure*}

\subsection{Auxiliary task}
\label{sec:AuxTask}

The network for the auxiliary task is shown at the top of
Figure~\ref{fig:MTTDSCnetwork0}.
The auxiliary model consists of a
left-to-right \GRUauxLR, a right-to-left \GRUauxRL,
and a fully-connected layer $\Waux \in \R^{2D\times 3}$.
The auxiliary model is trained with 
tweets that are accompanied by whole-tweet sentiment labels
from $\{-1,0,1\}$.  
First \GRUauxLR\ and \GRUauxRL\ are applied
over the entire tweet (positions $1,\ldots,N$).
At every token position $i$,
we construct the concatenation
\begin{align*}
\bigl[ \GRUauxLR[i-1], \GRUauxRL[i+1] \bigr]
\end{align*}
These are then averaged over positions to get a fixed-size
pooled representation $\bar{\pmb{x}} \in \R^{2D}$ of the whole tweet:
\begin{align}
\bar{\pmb{x}} &=
\frac{1}{N}\sum_{i=1}^N
\bigl[ \GRUauxLR[i-1], \GRUauxRL[i+1] \bigr]
\end{align}
Average pooling lets the auxiliary model learn useful features at all positions of the tweet.  This helps the primary task, as the target entity of the primary task can be at any position.  The whole-tweet prediction output is $\softmax(\bar{\pmb{x}}\, \Waux) \in \Delta^3$.  Again, cross-entropy loss is used.

\subsection{Main task}
\label{sec:MainTask}

Beyond the auxiliary model components, the main task uses a left-to-right \GRUmainLR, a right-to-left \GRUmainRL, and a fully connected layer $\Wmain \in \R^{4D\times 3}$ as model components.

Let target entity be at token position $i$.
\GRUauxLR\ and \GRUmainLR\ are run over positions $1,\ldots,i-1$.
\GRUauxRL\ and \GRUmainRL\ are run over positions $i+1,\ldots,N$.
The four resulting state vectors $\GRUauxLR[i-1]$,
$\GRUauxRL[i+1]$, $\GRUmainLR[i-1]$, and $\GRUmainRL[i+1]$
are concatenated into a vector in $\R^{4D}$, which is input into
the fully-connected layer followed by a softmax.
\begin{align}
\softmax\Bigl(\bigl[ 
\GRUauxLR[i-1], \GRUmainLR[i-1], 
\GRUauxRL[i+1], \GRUmainRL[i+1] \bigr] \Wmain \Bigr).
\end{align}
Our network for the situation where the auxiliary and main tasks do not share instances is shown in Figure~\ref{fig:MTTDSCnetwork0}.

\subsection{Training the tasks}
\label{sec:TwoTasks}

Suppose the auxiliary task has instances $\{(x_i,y_i): i=1,\ldots,A\}$ and the main task has instances $\{(x_j,y_j): j=1,\ldots,M\}$.  Let $\text{GRU}_*^*$ be all the GRU model parameters in $\{\GRUauxLR, \GRUauxRL, \GRUmainLR, \GRUmainRL\}$.  Then our overall loss objective is 
\begin{multline}
\sum_{i=1}^A \text{loss}_\textsc{aux}(x_i, y_i; \GRUauxLR, \GRUauxRL, \Waux)
%\\[-3ex] 
+ \alpha
\sum_{j=1}^M \text{loss}_\textsc{main}(x_j, y_j; \text{GRU}^*_*, \Wmain)
\end{multline}
Standard cross-entropy is used for both $\text{loss}_\textsc{aux}$ and $\text{loss}_\textsc{main}$.  Before training the full objective above, we pre-train only the auxiliary task for one epoch.  
The situation where instances may be shared between the auxiliary and main tasks is similar, except that GRU cells are now directly shared between auxiliary and main tasks.
We anticipate this multi-task setup to do better than, say, fine-tuning word embeddings in TDLSTM, because the auxiliary task is better related to the main task than unsupervised word embeddings.  By the same token, we do not necessarily expect our auxiliary learner to outperform more direct approaches for the auxiliary task --- its goal is to supply better word/span representations to the main task.

\subsection{Implementation details}
\label{sec:SystemDetails}

\noindent\textbf{\underline{GRU instead of LSTM}:}
We used GRUs instead of LSTMs, which are more common in prior work.  GRUs have fewer parameters and are less prone to overfitting.  In fact, our TDGRU replacement performs better than TDLSTM (Tables \ref{table:res} and \ref{table:ourData}).

\noindent\textbf{\underline{Hyperparameters}:}
We set the hidden unit size of each GRU network as 64.
Recurrent dropout probability of the GRU is taken as 0.2.
We also used a dropout of 0.2 before the last fully connected layer. 
For training the models we used the Adam optimizer with learning rate of 0.001, $\beta_1=0.9$, $\beta_2=0.999$.  We used a mini-batch size of 64.

\noindent\textbf{\underline{Ensemble}:} While reimplementing and/or running baseline system codes, we saw large variance in test accuracy scores for random initializations of the network weights. We improved the robustness of our networks by using an ensemble of the same model trained on the complete dataset with different weight initializations.  The output class scores of the final model are the average of the probabilities returned by members of the ensemble. For a fair comparison, we also use the same ensembling for all our baselines.

\noindent\textbf{\underline{Word embeddings}:} MTTDSC, TDLSTM and TCLSTM use GloVe embeddings \citep{PenningtonSM2014GloVe} trained on the Twitter corpus.  %To remain consistent with published numbers, TDParse uses SSWE embeddings, %\citep{TangWYZLQ2014SSWE},
% which may give it some benefit compared to using GloVe embeddings.

\section{Experiments}
\label{sec:Expt}

We summarize datasets, competing approaches, and accuracy measures, followed by a detailed performance comparison and analysis.

\subsection{Datasets for auxiliary SC task} 

\textbf{\underline{Go \protect\citep{go2009twitter}:}} 
This is a whole-passage SC dataset, containing 1.6M tweets automatically annotated using emoticons, highlighting that SC labeling can be easier to acquire at large scale.  It has only positive and negative classes.

\noindent\textbf{\underline{Sanders~\protect\citep{Sanders2011TwitterSentiment}:}}
The second dataset is provided by Sanders Analytics and has 5,513 tweets over all 3 classes.  These are manually annotated.

\subsection{Datasets for main TDSC task} \label{sec:maindata}

\textbf{\underline{Dong~\protect\citep{DongWTTZX2014AdaRnnTDSC}:}}
Target entities are marked in tweets (one target entity per tweet), and one of three sentiment labels manually associated with each target.  The training and test folds contain 6,248 and 692 instances respectively.  The class distribution is 25\%, 50\% and 25\% for negative, neutral and positive respectively.

\noindent\textbf{\underline{Election1~\protect\citep{WangLZP2017TDParse}:}}
Derived from tweets about the recent UK election, this dataset contains 3,210 training tweets that mention 9,912 target entities and 867 testing tweets that mention 2,675 target entities.  The class distribution is 45.3\%, 36.5\%, and 17.7\% for negative, neutral and positive respectively, which is highly unbalanced.  There are an average of 3.16 target entities per tweet.

\noindent\textbf{\underline{Election2:}}
In this paper, we introduce a new TDSC dataset, also based on UK election tweets.
%\footnote{We took UK election tweets for two reasons: (i)~we found the annotation of {\bf Election1} inconsistent and biased, (ii)~the required number of person-specific named entities was not sufficient in {\bf Election1} to be used for anomaly detection.}.
We first curated a list of candidate hashtags related to the UK General Elections, such as \#GE2017, \#GeneralElection and \#VoteLabour.  
%Using Twitter's streaming API, we collected tweets which contain at least one of these hashtags.  
The collection was done during a period of 12 days, from June 2, 2017 through June 14, 2017.  After removing retweets and duplicates (tweets with the same text), we ended up with 563,812 tweets.  
%All non-ASCII characters were removed from the tweets. 
% They were tagged for named entities using a NER model trained specifically for tweets, developed by \citet{RitterCE2011NerTweet}. 
After running the named entity tagger, we observed that 158,978 tweets (28.19\%) had at least one named entity, 38,809 tweets (6.88\%) had at least two named entities and the remaining 7,992 tweets (1.42\%) had three or more named entities.
We took all the tweets which had at least two named entities, and randomly sampled an equal number of tweets from the set of tweets which had only one named entity. %This was done so as to ensure that the final dataset is rich in entities.  From this combined set, 6,699 tweets were randomly sampled and annotated manually by annotators for target specific sentiment, following the exact guidelines and the annotation system developed by \cite{WangLZP2017TDParse}.  1,836 (19.13\%) were marked as positive, 5,024 (52.35\%) were marked as neutral and 2,737 (28.52 \%) were marked as negative by the annotators. 

% Both Election1 and Election2 are heavily skewed in favor of label~0 and against label~$+1$, making macro-averaged F\textsubscript{1} score more meaningful than micro-averaged accuracy.

% In all there were 9,597 entities.

\begin{table*}[t]
\caption{Performance of various methods on {\bf Dong} dataset: (a)~overall, (b)~class-wise. MTTDSC beats other baselines across diverse performance measures.}
\label{table:res}
\centering
\scalebox{0.9}{
\begin{tabular}{|c|c|c|c|c|c|c|c||c|p{0.04mm}|c|c|c|}
 \multicolumn{1}{c}{}   & \multicolumn{8}{c}{(a)}&\multicolumn{1}{}{} &\multicolumn{3}{c}{(b)}\\    
    \cline{1-9} \cline{11-13}
    
    & \multirow{2}{*}{\textbf{Model}} & \multicolumn{6}{c||}{\textbf{3 Class Performance}}  & \multirow{1}{*}{\textbf{2-class}} & & \multicolumn{3}{|c|}{{\bf Class-wise F\textsubscript{1}}} \\ \cline{3-8} \cline{11-13}
    &                       & \textbf{Accuracy} & \textbf{Precision} & \textbf{Recall} & \textbf{F\textsubscript{1}} & \textbf{MAE} & \textbf{PIR\%} & {\bf F\textsubscript{1}}   & & $\mathbf{-1}$ & ${\bf 0}$ & ${\bf +1}$  \\ \cline{1-9} \cline{11-13}
  \parbox[t]{3mm}{\multirow{11}{*}{\rotatebox[origin=c]{90}{{\bf Baseline}}}}  & LSTM & 66.5 & 66.0 & 63.3 & 64.7 & 0.367 & 5.39 & 60.6 & & 64.4 & 71.8 & 56.9\\
    & Target-ind & 67.3 & 69.2 & 61.7 & 63.8 & 0.351 & 5.45 & 58.8 & & 60.9 & 73.6 & 56.8 \\
    & Target-dep+ & 70.1 & 69.9 & 65.9 & 67.7 & 0.341 & 5.31 &  62.6  & & 65.9 & 75.7 & 59.4 \\
    \cline{2-9} \cline{11-13}
    &TDLSTM & 70.8 & 69.5 & 68.8 & 69.0 & 0.335 & 5.61 & 65.7& &  68.9 &75.6 & 62.5\\
    &TCLSTM & 71.5 & 70.3 & 69.4 & 69.5 & 0.321 & 5.47 & 67.2 & & 68.4 & 75.1 & 65.9 \\
    &SLUICE & 69.2 & 67.9 & 67.2 & 67.5 & 0.348 & 5.90 & 64.4 & & 67.6 & 73.7 & 61.6\\
    &Naive-Seg+ & 70.7 & 70.6 & 65.9 & 67.7 & 0.332 & 5.54 & 63.2 & & 66.0 & 76.3 & 60.0 \\
    &TDParse &  71.0 & 70.5 & 67.1 & 68.4 & 0.331 & 5.86 & 64.3 & & 65.8 & 76.7 & 62.7  \\
    &TDParse+ & 72.1 & 72.2 & 68.3 & 69.8 & 0.312 & 5.00 & 66.0 & & 67.5 & 77.3 & 64.5 \\
     &TDParse+ (m) & 72.5 & 72.6 & 68.9 & 70.3 & 0.308 & 4.95 & 66.6 & & 68.3 & 77.6 & 65.0\\
    \cline{1-9} \cline{11-13}
    
\parbox[t]{3mm}{\multirow{5}{*}{\rotatebox[origin=c]{90}{{\bf Our}}}}	
    &TDGRU & 71.0 & 70.2 & 68.7 & 69.3 & 0.324 & 5.14 & 66.4 & & 67.8 & 75.1 & 65.0\\
    &TDGRU+SVM & 71.7 & 71.4 & 68.7 & 69.7 & 0.315 & 5.16 & 66.5  & & 68.9 & 76.2 & 64.1\\
    &TD naive MTL & 63.0 & 63.2 & 57.4 & 59.1 & 0.403 & 6.22  & 46.2 & &55.0 & 70.3 & 52.0\\
    &TDFT & 73.3 & 72.7 & 70.8 & 71.6 & 0.299 & 5.17 & 68.8 & & 70.0 & 77.4 & \textbf{67.7} \\
    &\textbf{MTTDSC} & \textbf{74.1} & \textbf{74.0} & \textbf{71.7} & \textbf{72.7} & \textbf{0.286} & \textbf{4.22} & \textbf{70.0} & & {\bf 72.8} & {\bf 77.9} & {67.3}\\ \cline{1-9} \cline{11-13}
    \end{tabular}
}
\end{table*}

\begin{table*}[th]
\caption{Performance of various methods on our \textbf{Election2} dataset: (a)~overall, (b)~class-wise.  The extreme label skew makes it easy for simpler algorithms to do well at MAE and PIR, although MTTDSC still leads in traditional measures, and recognizes neutral content better.}
\label{table:ourData}
\scalebox{0.9}{
\begin{tabular}{|c|c|c|c|c|c|c|c||c|p{0.05mm}|c|c|c|}
 \multicolumn{1}{c}{}   & \multicolumn{8}{c}{(a)}&\multicolumn{1}{}{} &\multicolumn{3}{c}{(b)}\\    
    \cline{1-9} \cline{11-13}
    & \multirow{2}{*}{\textbf{Model}} & \multicolumn{6}{c||}{\textbf{3 Class Performance}}  & \multirow{1}{*}{\textbf{2-class}} & & \multicolumn{3}{|c|}{{\bf Class-wise F\textsubscript{1}}} \\ \cline{3-8} \cline{11-13}
    &                       & \textbf{Accuracy} & \textbf{Precision} & \textbf{Recall} & \textbf{F\textsubscript{1}} & \textbf{MAE} & \textbf{PIR\%} & {\bf F\textsubscript{1}}   & & $\mathbf{-1}$ & ${\bf 0}$ & ${\bf +1}$  \\ \cline{1-9} \cline{11-13}
  \parbox[t]{3mm}{\multirow{11}{*}{\rotatebox[origin=c]{90}{{\bf Baseline}}}}  & LSTM &55.6 & 53.4 & 43.1 & 42.3 & 0.484 & 7.76 & 29.0& & 47.9& 57.2 & 56.8\\
    & Target-ind & 55.6 & 52.3 & 43.7 & 42.7 & 0.484 & 7.87 & 30.0 & & 42.5 & 68.1 & 17.4 \\
    & Target-dep+ & 59.4 & 59.6 & 48.1 & 48.4 & 0.447 & 7.30 & 37.2 & & 47.2 & 70.9 & 27.1 \\ \cline{2-9} \cline{11-13}   
    &TDLSTM & 58.6 & 54.5 & 50.0 & 50.7 & 0.479 & 10.46 & 41.1 &  & 48.0 & 70.0 & 34.2 \\
    &TCLSTM & 58.4 & 54.3 & 49.3 & 50.1 & 0.474 & 9.73 & 40.1 & & 45.1 & 70.1 & 35.1 \\
    &SLUICE & 58.8 & 54.8 & 53.0 & 52.9 & 0.489 & 10.9 & 45.2 &  & 54.9 & 68.3 & 35.4 \\
    &Naive-Seg+ & 60.5 & 60.0 & 51.1 & 52.2 & \textbf{0.433} & \textbf{7.06} & 43.0 & & 51.1 & 70.5 & 34.9 \\
    &TDParse & 58.9 & 56.5 & 50.8 & 51.6 & 0.460 & 8.67 & 42.9 & & 51.3 & 68.8 & 34.5 \\
    &TDParse+ & 61.1 & 59.8 & 52.7 & 53.6 & 0.438 & 8.16 & 45.1 & & \textbf{53.4} & 70.8 & 36.7 \\
     &TDParse+(m) & 60.6 & 59.1 & 52.2 & 53.1 & 0.440 & 8.40 & 44.5 & & 52.4 & 70.5 & 36.6 \\
    \cline{1-9} \cline{11-13}
    
\parbox[t]{3mm}{\multirow{5}{*}{\rotatebox[origin=c]{90}{{\bf Our}}}}	
    &TDGRU & 59.5 & 55.4 & 53.0 & 53.8 & 0.475 & 10.67 & \textbf{45.8} & & 50.8 & 69.7 & \textbf{40.8} \\
    &TDGRU+SVM & 59.2 & 56.0 & 51.6 & 52.5 & 0.469 & 9.93 & 43.9 & & 50.0 & 69.6 & 37.9\\
    &TD naive MTL &  56.5 & 57.1 & 43.2 & 41.4 & 0.473 & 7.04& 27.3 & & 51.2 &57.5 & 62.7\\
    &\textbf{MTTDSC} & {\bf 61.6} & {\bf 60.1} & {\bf 53.1} & {\bf 54.1} & 0.439 & 8.79 & 45.3 & & 52.6 & {\bf 71.8} & 38.0 \\ \cline{1-9} \cline{11-13}
    \end{tabular}
    }
\end{table*}
\vspace{-5mm}

\subsection{Details of performance measures}
% \vspace{-2mm}

Past TDSC work reports on 0/1~accuracy and macro averaged F\textsubscript{1} scores, and we do so too, for 3-class $\{-1,0,1\}$ instances and 2-class $\{-1,1\}$ subsets.
However, SC is fundamentally a regression or ordinal regression task; e.g., $\text{loss}(-1,1) > \text{loss}(-1,0) > \text{loss}(-1,-1)=0$.  Evaluating ordinal regression in the face of class imbalance can be tricky \citep{BaccianellaES2009OReval}.  In addition, the system label may be discrete from $\{-1,0,1\}$ or continuous in $[-1,1]$.  Therefore we report on two additional performance measures.  Let $(x_i,y_i, \hat{y}_i)$ be the $i$th of $I$ instances, comprised of a tweet, gold label, and system-estimated label.

\textbf{Mean absolute error (MAE)}
It is defined as $(1/I)\sum_i |y_i - \hat{y}_i|$.  Downstream applications that use the numerical values of $\hat{y}$ will want MAE to be small.

\textbf{Pair inversion rate (PIR):}
For a pair of instances $(i,j)$, if $y_i > y_j$ but $\hat{y}_i < \hat{y}_j$, that is an \emph{inversion}.  PIR is then defined as $\left(\sum_{i\ne j} 
\left\llbracket (y_i - y_j) (\hat{y}_i - \hat{y}_j) < 0 
\right\rrbracket \right)\left/\binom{I}{2}\right.$.  Closely related to the area under the curve (AUC) for 2-class problems, PIR is widely used in Information Retrieval \citep{Joachims2002ranksvm}.

\subsection{Various methods and their performance}
\label{sec:MethodsAndPerfs}

Table~\ref{table:res} shows aggregated and per-class accuracy for the competing methods.  It has three groups of rows.  The first group includes methods that use no or minimal target-specific processing.  The second group includes the best-known recent target-dependent methods.  The third group includes our methods and their variants, to help elucidate and justify the merits of our design.

\noindent \textbf{\underline{Target independent baselines}:}
In the first block, \textbf{LSTM} means a whole-tweet LSTM was applied, followed by a linear SVM on the final state.  \textbf{Target-ind} \citep{jiang2011target} pools embedding features from the entire tweet.  \textbf{Target-dep+} extends Target-dep. %\todo{Target-dep+ is not target independent, change subsection} 
\textbf{Target-dep+} extends Target-dep by identifying sentiment-revealing context features with the help of multiple lexicons (such as SentiWordNet \footnote{\protect\url{http://sentiwordnet.isti.cnr.it/}}).

\noindent \textbf{\underline{Prior target-dependent baselines}:}
The second block shows more competitive target-aware TDSC approaches.
\textbf{TDLSTM} and \textbf{TCLSTM} are from \citet{TangQFL2015TDLSTM}.
\textbf{Naive-Seg+} segments the tweet using punctuations. 
Word vectors in each segment are pooled to give a segment embedding.
Additional features are generated from the left and right contexts based on
multiple sentiment lexicons.  \textbf{TDParse} \citep{WangLZP2017TDParse} uses a syntactic parse to pool embeddings of words connected to the target.  \textbf{TDParse+} extends TDParse by adding features from sentiment lexicons.  \textbf{TDParse+(m)} considers the presence of the same target multiple times in the tweet.  Feature vectors generated from multiple target positions are merged using pooling functions.

\noindent \textbf{\underline{{MTTDSC and variations:}}}
The third block shows \textbf{MTTDSC} and some variations.
\textbf{TDGRU} replaces the two LSTMs of TDLSTM with two GRUs \citep{ChungGCB2014GRUvsLSTM} which have fewer parameters but perform slightly better than LSTMs.  In \textbf{TDGRU+SVM}, we first train the TDGRU model.
Then, at entity position $i$, we extract the features $\bigl[\text{GRU}_\textsc{lr}[i]; \text{GRU}_\textsc{rl}[i]\bigr]$ of the two GRU models and train an SVM with RBF kernel with the extracted features.  TDGRU with SVM is expected to perform better due to the non-linear nature of the features at the penultimate layer, which the SVM can then recognize without overfitting problems.
\textbf{TD~naive~MTL} is similar to MTTDSC, but, rather than having separate GRUs for primary and auxiliary tasks, shared $\text{GRU}_\textsc{lr}$ and $\text{GRU}_\textsc{rl}$ are used for both tasks.  The tasks are trained jointly as in MTTDSC.  In \textbf{TDFT}, we first train $\text{GRU}_\textsc{lr}$, $\text{GRU}_\textsc{rl}$ and $W_\textsc{aux}$ on the auxiliary whole-passage SC task.  We then use the weights of $\text{GRU}_\textsc{lr}$ and $\text{GRU}_\textsc{rl}$ learnt by the auxiliary task in TDGRU and train it on TDSC with a new $W_\textsc{main}$.

\noindent \textbf{\underline{Observations:}}
Table \ref{table:res} shows that MTTDSC outperforms all the baselines across all the measures on {\bf Dong} dataset.
MTTDSC achieves 2.8\%, 3.41\%, 7.14\% and 24.5\% relative improvements
in accuracy, F\textsubscript{1}, MAE and PIR respectively over 
TDParse+(m) (best model by \citet{WangLZP2017TDParse}).
The improvement in 2-class F\textsubscript{1} is also substantial (5.1\%).
MTTDSC maintains a better balance between precision, recall
and F\textsubscript{1} across the three classes (Table \ref{table:res}(b)).

\begin{table}[t]
\caption{Performance comparison on {\bf Election1} dataset. The results of the baselines are taken from Table~3 of \citet{WangLZP2017TDParse} where TDPWindow-12 (which extracts features exactly like TDParse+, but limits the size of left and right contexts to 12 tokens) was reported as the best model.  To save space, we report the accuracy w.r.t. only three measures.  The broad trends are similar to Election2.} \label{table:tdparse}
\centering
\begin{tabular}{|c|c|c|c|}
\hline
{\bf Model} & {\bf Accuracy} & {\bf 3-Class F\textsubscript{1}} & {\bf 2-class F\textsubscript{1}}\\
\hline
Target-ind & 52.30 & 42.19 & 40.50  \\
Target-dep+ & 55.85 & 43.40 & 40.85  \\
TDParse & 56.45 & 46.09 & 43.43 \\
TDPWindow-12 & {\bf 56.82} & 45.45 & 42.69 \\
\hline
TDGRU   & 55.46  & 47.12      & 44.22 \\
MTTDSC  & 56.67  & {\bf 47.71} & {\bf 45.58} \\
\hline
\end{tabular}
\end{table}

TDFT improves on TDLSTM and TDGRU because it learns important features during pre-training.  TDFT is better than TD~naive~MTL; jointly training the latter results in auxiliary loss prevailing over primary loss.  TD~naive~MTL also loses to MTTDSC, because, in fine tuning, the primary task training makes the model forget some auxiliary features critical for the primary task.  Summarizing, MTTDSC's gains are not explained by the large volume of auxiliary data alone; good network design is critical.

Table~\ref{table:ourData} shows results for \textbf{Election2}.  The trend is preserved, with our gains in macro-F\textsubscript{1} being more noticeable than micro-accuracy.  This is expected given the label skew.
%\todo{Why no MAE or PIR here? - some models not implemented , directly coptied the results}
Table~\ref{table:tdparse} shows similar behavior on the \textbf{Election1} dataset.  Although TDPWindow-12 is slightly better for 0/1 accuracy, MTTDSC achieves 4.97\% and 6.77\% larger F\textsubscript{1} score for 3-class and 2-class sentiment classification respectively.

\begin{table*}[th]
\caption{Word sensitivity studies.  (Must be viewed in color.)
Green words are regarded as positive and red words are regarded as
negative by the respective RNNs.  Intensity of color roughly represents magnitude of sensitivity.  TDLSTM makes mistakes in estimating the polarity of words independent of context, which lead to incorrect predictions.  Assisted by the auxiliary task, MTTDSC avoids such mistakes.}
\label{fig:heatmaps}
\centering  \includegraphics[width=1\textwidth]{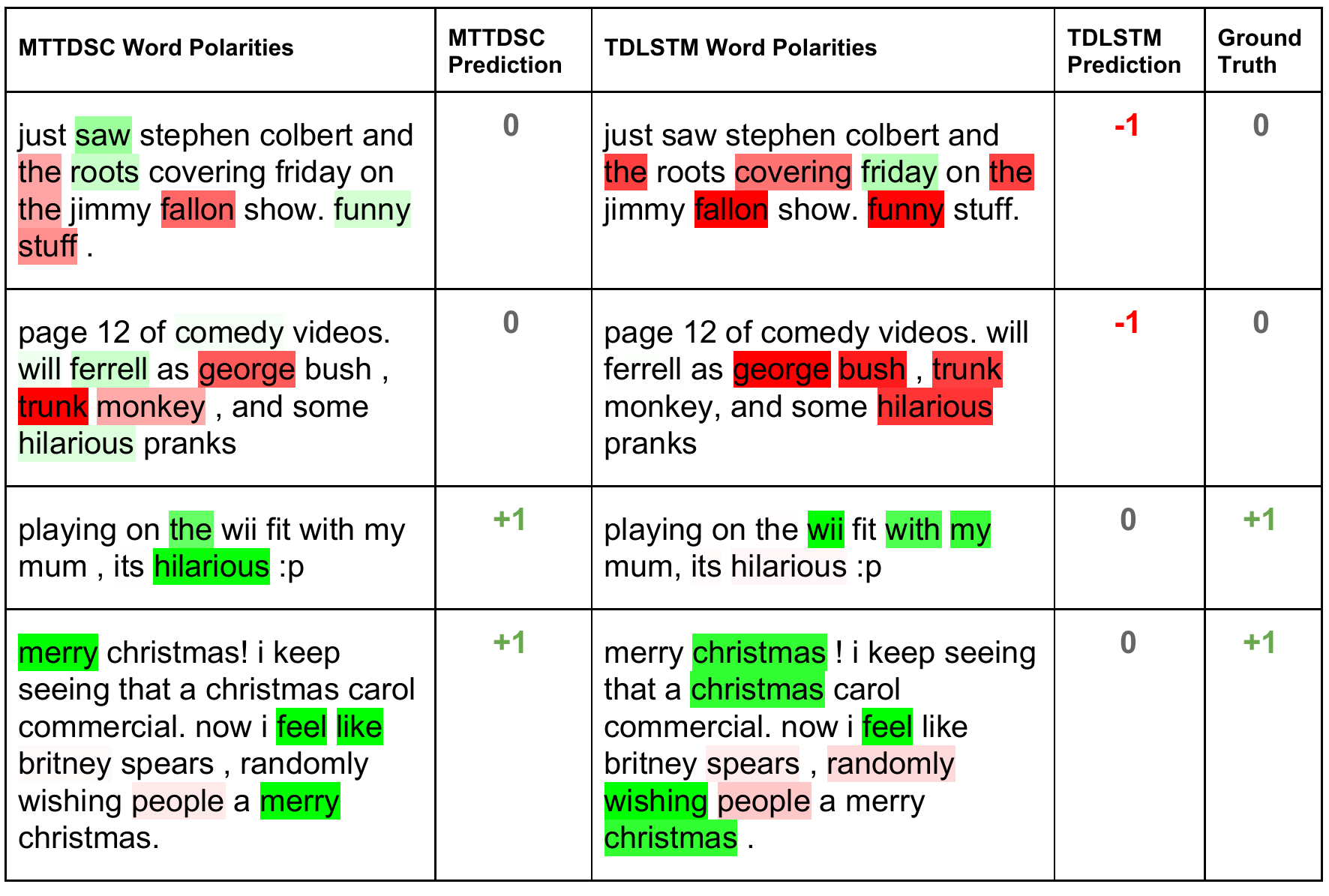}
\end{table*}

\vspace{-5mm}

\subsection{Side-by-side diagnostics and anecdotes}
\label{sec:anecdotes}

Given their related architectures, we picked TDLSTM and MTTDSC, and 
focused on instances where MTTDSC performed better than TDLSTM, to tease out the reasons for the improvement.

\noindent\textbf{\underline{Word-level sensitivity analysis}:}
For each word in the context of the target entity, we replaced the word with \texttt{UNK} (unknown word) and noted the drops in scores of labels $+1$ and~$-1$. 
A large drop in the score of label~$+1$ means the word was regarded as strongly positive, and a large drop in the score of label~$-1$ means the word was regarded as strongly negative.  We use these scores to color-code context words in the form of a heatmap.
Figure~\ref{fig:heatmaps} shows the positive and the negative words highlighted accordingly to their sensitivity scores. The words highlighted in green color contribute to the positive label and the words highlighted red contribute to the negative label.
In the first row, MTTDSC correctly identifies \emph{funny} as a positive word, whereas TDLSTM considers \emph{funny} to be a negative word.  TDLSTM also finds stronger negative polarity in neutral words like \emph{the} and \emph{covering}.
In the second row, MTTDSC correctly identifies \emph{hilarious} as a positive word, whereas TDLSTM finds \emph{hilarious} strongly negative.  
In the third row, MTTDSC finds \emph{hilarious} positive, whereas TDLSTM misses the signal.  Although TDLSTM correctly identifies more positive words in the fourth row than MTTDSC, it also incorrectly identifies negative words like \emph{randomly} and \emph{people}, leading to an overall incorrect neutral prediction.
The examples show that TDLSTM either misses or misclassifies crucial emotive, polarized context words.  % Of all context words that are also found in SentiWordNet, we notice almost 6\% misclassified by TDLSTM, despite using pre-trained word embeddings.

\section{Conclusion}
\label{sec:End}

We presented MTTDSC, a multi-task system for target-dependent sentiment classification. By exploiting the easier auxiliary task of whole-passage sentiment classification, MTTDSC improves on recent TDSC baselines.  The auxiliary LSTM learns to identify corpus-specific, position-independent sentiment in words and phrases, whereas the main LSTM learns how to associate these sentiments with designated targets.  We tested our model on three benchmark datasets, of which we introduce one here, and obtained clear gains in accuracy compared to many state-of-the-art models.  % Simpler, generic MTL approaches fail to get the same benefits.  MTTDSC also provides more interpretable and intuitive labeling decisions.  

\section{Acknowledgement}
The project was partially supported by IBM,  Early Career Research Award (SERB, India), and the Center for AI, IIIT Delhi, India. 

\newpage
%\input{anomaly.tex}
%We also showed the efficacy of MTTDSC through the extrinsic task of anomaly detection. 

\input{anomaly}

\bibliographystyle{IEEEtran}
\bibliography{mttdsc,voila}

% \begin{thebibliography}{00}
% \bibitem{b1} G. Eason, B. Noble, and I. N. Sneddon, ``On certain integrals of Lipschitz-Hankel type involving products of Bessel functions,'' Phil. Trans. Roy. Soc. London, vol. A247, pp. 529--551, April 1955.
% \bibitem{b2} J. Clerk Maxwell, A Treatise on Electricity and Magnetism, 3rd ed., vol. 2. Oxford: Clarendon, 1892, pp.68--73.
% \bibitem{b3} I. S. Jacobs and C. P. Bean, ``Fine particles, thin films and exchange anisotropy,'' in Magnetism, vol. III, G. T. Rado and H. Suhl, Eds. New York: Academic, 1963, pp. 271--350.
% \bibitem{b4} K. Elissa, ``Title of paper if known,'' unpublished.
% \bibitem{b5} R. Nicole, ``Title of paper with only first word capitalized,'' J. Name Stand. Abbrev., in press.
% \bibitem{b6} Y. Yorozu, M. Hirano, K. Oka, and Y. Tagawa, ``Electron spectroscopy studies on magneto-optical media and plastic substrate interface,'' IEEE Transl. J. Magn. Japan, vol. 2, pp. 740--741, August 1987 [Digests 9th Annual Conf. Magnetics Japan, p. 301, 1982].
% \bibitem{b7} M. Young, The Technical Writer's Handbook. Mill Valley, CA: University Science, 1989.
% \end{thebibliography}

\end{document}